\documentclass[10pt, a4paper]{article}

\usepackage{lrec2026} 
\usepackage{booktabs}
\usepackage{times}
\usepackage{latexsym}
\usepackage{enumitem}

 \newenvironment{tight_itemize}{
    \begin{itemize}[noitemsep, topsep=0pt, parsep=0pt, partopsep=0pt, leftmargin=*, before=\vspace{5pt}, itemsep=5pt, labelsep=0.5em]
    \setlength{\parindent}{0pt}
}{
    \end{itemize}
}

\title{SACRED: A Faithful Annotated Multimedia Multimodal Multilingual Dataset for Classifying Connectedness Types in Online Spirituality}

\name{
{\large\bfseries
\begin{tabular}{c}
Qinghao Guan$^{1}$, Yuchen Pan$^{2}$, Donghao Li$^{2}$, Zishi Zhang$^{2}$, Yiyang Chen$^{2}$, \\ Lu Li$^{3,4}$, Flaminia Canu$^{1}$, Emilia Volkart$^{1}$, Gerold Schneider$^{2}$
\end{tabular}
}
}

\address{
$^{1}$Department of Communication and Media Research, University of Zurich, Zurich, Switzerland \\
$^{2}$Department of Computational Linguistics, University of Zurich, Zurich, Switzerland \\
$^{3}$Department of Informatics, University of Zurich, Zurich, Switzerland \\
$^{4}$ Ping An Technology, Shanghai, China
}
         
\abstract{
In religion and theology studies, spirituality has garnered significant research attention for the reason that it not only transcends culture but offers unique experience to each individual. However, social scientists often rely on limited datasets, which are basically unavailable online. In this study, we collaborated with social scientists to develop a high-quality multimedia multi-modal datasets, \textbf{SACRED}, in which the faithfulness of classification is guaranteed. Using \textbf{SACRED}, we evaluated the performance of 13 popular LLMs as well as traditional rule-based and fine-tuned approaches. The result suggests DeepSeek-V3 model performs well in classifying such abstract concepts (i.e., 79.19\% accuracy in the Quora test set), and the GPT-4o-mini model surpassed the other models in the vision tasks (63.99\% F1 score). Purportedly, this is the first annotated multi-modal dataset from online spirituality communication. Our study also found a new type of connectedness which is valuable for communication science studies. 
 \\ \newline  }

\begin{document}

\maketitleabstract

\section{Introduction}
From the primal reverence of natural forces and totems in ancient times to the complex theological systems today, spirituality has been an inextricable part of human society. The concept of totems, for instance, exemplifies the enduring spirituality pursuit throughout history. Rooted in the earliest human societies, totemism reflects a profound connection between people and the natural world, symbolizing a relationship that is both sacred and essential for the cultural identity of indigenous communities. These totems, be they animals, plants, or celestial bodies \citep{Bolatova2019} were not mere symbols but represented a deep, spiritual kinship and understanding of the interconnectedness of all life. Therefore, human spirituality or religion are considered indispensable to human functioning and survival \citep{Lee2015}.

Spirituality holds that all life is interconnected \citep{Spaniol2002}. At its core, relationality constitutes the essence of spiritual belief \citep{deSouza2003, deSouza2004, deSouza2012}. It can be argued that the impulse to connect represents one of humanity's most fundamental and inherent desires. The ability to forge spiritual connections -- whether with the divine, within oneself, with others, or with the natural world -- plays a crucial role in shaping individual identities and in defining our roles within the world.

\begin{figure}[t]
    \centering
    \includegraphics[width=1\linewidth]{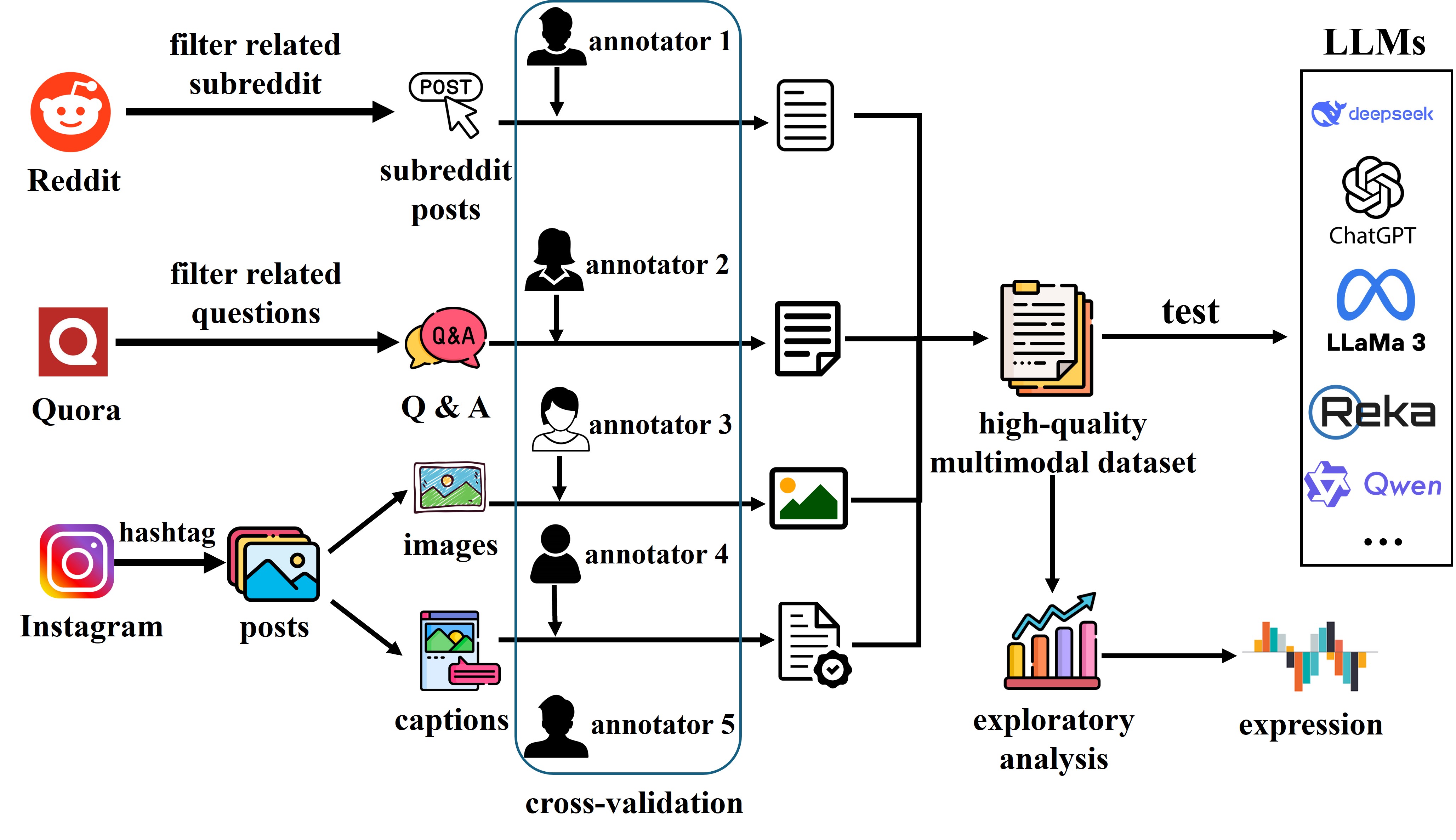}
    \caption{Workflow Diagram of SACRED Dataset Annotation and LLMs Evaluation}
    \label{fig:enter-label}
\end{figure}

\begin{table*}[!ht]
\centering
\caption{Concepts of Connectedness and Examples}
\begin{tabular}{>{\centering\arraybackslash}m{3cm} >{\centering\arraybackslash}m{5cm} >{\centering\arraybackslash}m{6.5cm}}

\toprule
\textbf{Concept} & \textbf{Description} & \textbf{Example} \\
\midrule
Connectedness to Self & authenticity, inner harmony and inner peace, consciousness, self-knowledge, and experiencing and searching for meaning in life & Just spent the morning in deep meditation and self-reflection. Amazing how connecting with my inner self brings clarity and peace to my spirit. \#InnerPeace \#SelfAwareness \\
\addlinespace
Connectedness to Others & compassion, caring, gratitude, and wonder & Just returned from volunteering at the local shelter, and my spirit feels so enriched! Connecting with others through acts of kindness truly uplifts the soul. \#CommunityLove \#MentalWellbeing \\
\addlinespace
Connectedness to Nature & the deep sense of relationship that individuals feel with the natural world and understanding of humanity's place within the broader ecological system & Hiking through the forest today, I felt an incredible connection to the earth. The serenity of nature rejuvenates my spirit and calms my mind. \#NatureHealing \#SoulConnection \#EarthSpirit \\
\addlinespace
Connectedness to Transcendence & something or someone beyond the human level, such as the universe, transcendent reality, a higher power, or God & Attended a beautiful church service this morning. As we sang together and lifted our voices in prayer, I felt an incredible connection to God that filled my spirit with joy and hope. Truly a transcendent experience. \#Faith \#DivineConnection \#SpiritualUpliftment \\
\bottomrule
\label{tab:connectedness}
\end{tabular}
\end{table*}

Although very popular with audiences \citep[p.47]{Ramasubramanian2014}, spirituality in the context of social media is still considered to be an understudied area of research \citep{Janicke-Raney-2016}. Our paper makes the following contributions:

\begin{tight_itemize}
    \item This paper introduces \textbf{SACRED}, the first multimodal annotated dataset specifically designed for online spirituality research.
    \item The paper evaluates the visual and textual classification ability of 11 popular LLMs, including LLaMA3.1, Qwen2.5-VL, GPT-4o-mini, GPT-4o, DeepSeek-V3, GPT-4.1-mini, Claude Sonnet-4.5, Gemini-2.5-Flash, Reka, Gemma3, and LLaVA-1.5.
    \item This study, with the assistance of LLMs, uncovers a new type of connectedness within online spirituality and LLMs' understanding of spirituality. Our codes are available on GitHub\footnote{\url{https://github.com/Qinghao-Guan/Spirituality_LLMs4SSH-2026}}.
    
\end{tight_itemize}

\section{Theoretical Background}
\subsection{Online Spirituality}
Since the 1980s, the practice of religion in online settings have consistently grown. Initially marked by the creation of religious sub-groups on platforms like Usenet and through email, a variety of religious activities started to surface, drawing attention from both the media and academic circles \citep{Campbell2011}. Examples of this include the establishment of virtual places of worship and online sites for spiritual pilgrimages \citep{Campbell2006}. By the mid-1990s, academics began to seriously study these distinctive online socio-spiritual practices and contemplate the implications of transferring traditional religious beliefs and rituals to the digital realm. Over the last decade, even more creative and unique forms of online religious expression have emerged, ranging from religious "podcasting" \citep{Campbell2015} to religion-online \citep{Cheong2009} where served doctrinal interpretative and communal integrative functions are accomplished \citep{Helland2002}, and virtual worship spaces in Second Life for religious groups including Christians, Muslims, and Jews. 

\subsection{Dimensions of Connectedness}
Motivated by conceptual considerations \citep{Hill2000} and assessments of the significance of spirituality in the general population \citep{Skrzypinska2014}, much research focused on the multidimensional concepts of spirituality \citep{Demmrich_Huber2019}.

This research adopts the conceptualization of spirituality proposed by de Jager Meezenbroek et al. \citeyearpar{deJager2012}, who defined it as "one's striving for and experience of connection with oneself, others, nature and the transcendent". Their definition considered the spirituality as a universal human experience and emphasized the multidimensional nature of spirituality. Most of the current research, thus, is grounded in this comprehensive conceptualization.

Four dimensions of connectedness \citep{Ellison1983,Emmons2006} are often defined as follows: connectedness to self (involving authenticity, inner harmony/inner peace, consciousness, self-knowledge and experiencing and searching for meaning in life), connectedness to others (involving "compassion, caring, gratitude and wonder"), connectedness to Transcendence (referring to "something or someone beyond the human level"), and connectedness to nature (referring to the deep sense of relationship that individuals feel with the natural world and understanding of humanity's place within the broader ecological system). Examples are shown in Table \ref{tab:connectedness}.

\subsection{Definition of Spirituality}
\label{spirituality_definition}
A major challenge in researching internet-based religion is staying up-to-date with its swift evolution and transformations. This has posed a considerable problem in establishing theoretical models to study religious involvement on the World Wide Web \citep{Helland2005}. The accompanying issue is the definition of spirituality, which has been constantly changing from the early 20th century to the present \citep{Peng-Keller-spirituality}. Initially, spirituality was viewed primarily in terms of the sacred \citep{Otto1917,Eliade1959}. By 2014, the definition of spirituality had evolved to include both theistic and non-theistic dimensions \citep{Ramasubramanian2014}. This study considers both the theistic and non-theistic dimensions and use a strict definition of spirituality:

\begin{quote}
    Spirituality is the pursuit and practice of experiences and beliefs that influence and nurture the spirit, fostering personal growth, meaning, and a sense of connection to something greater than oneself.
\end{quote}

\section{Related Work}
\subsection{Textual Analysis of Online Spirituality}
Traditionally, textual data is analyzed by human coders who discern various characteristics and elements inherent in the text (i.g., content analysis in social science methodology). Manual annotation undoubtedly yields high-quality results. However, online platforms generate vast amounts of textual data on spirituality. Manually analyzing this data is impractical due to the overwhelming volume of data \cite{Hilbert-2019}, thus NLP techniques are essential tools for efficient analysis (cf. \citeauthor{Guan2025}, 2025 and \citeauthor{Guan-Lawi2024}, 2024). The prevalent techniques employed in analyzing posts related to spirituality are mainly term analysis, classification and topic modeling. Demmrich and Huber \citeyearpar{Demmrich_Huber2019} test six dimensions of Huber's model \citep{Huber2013} via frequency analysis and found that the "public practices" aspect, accounting for 11.69\% of the total coding, includes rites of passage, religious service, holiday celebrations, and public meditation. Additionally, machine learning models were used in this research field. Holmberg et al. \citeyearpar{Holmberg2016}, for instance, classified the tweets into 8 categories and concluded that these posts contain prayers that express praise, thanksgiving, devotion, care for other people, concern for one's own life and actions, questioning, despair, and even anger towards God. Sánchez-Garcés et al. \citeyearpar{Sanchez2021} employed a two-phase methodology (coding and sentiment analysis) to summarize key factors of patient interviews on their experiences during the disease. Their research revealed that patients often adopted a positive attitude (spiritual resilience) towards the symptoms and complications, significantly influenced by their religious faith. \citeauthor{Kim2020} (2020) utilized structural topic modeling (STM) to identify the latent dimensions of Religiosity/Spirituality, and they extracted three distinct topics, namely Experience of, Engagement in, and Essence of Transcendence. Similarly, \citeauthor{Winiger2025} (2025) presented a bottom-up approach to exploring the definitions of "religion" and "spirituality" by combining qualitative analysis with methods of distributional semantics. Their study demonstrates the potential of computational textual analysis for advancing research in religious studies. In the current LLM era, computational linguists (e.g., \citealp{Gao2025}), from an engineering perspective, developed a chat-based web interface called SpiritRAG, which is built on retrieval-augmented generation techniques. This tool, based on a database of 7,500 United Nations resolution documents addressing Religiosity/Spirituality issues, enables researchers and policymakers to efficiently search for information related to Religiosity/Spirituality.

\begin{figure*}[t]
    \centering
    \includegraphics[width=0.7\linewidth]{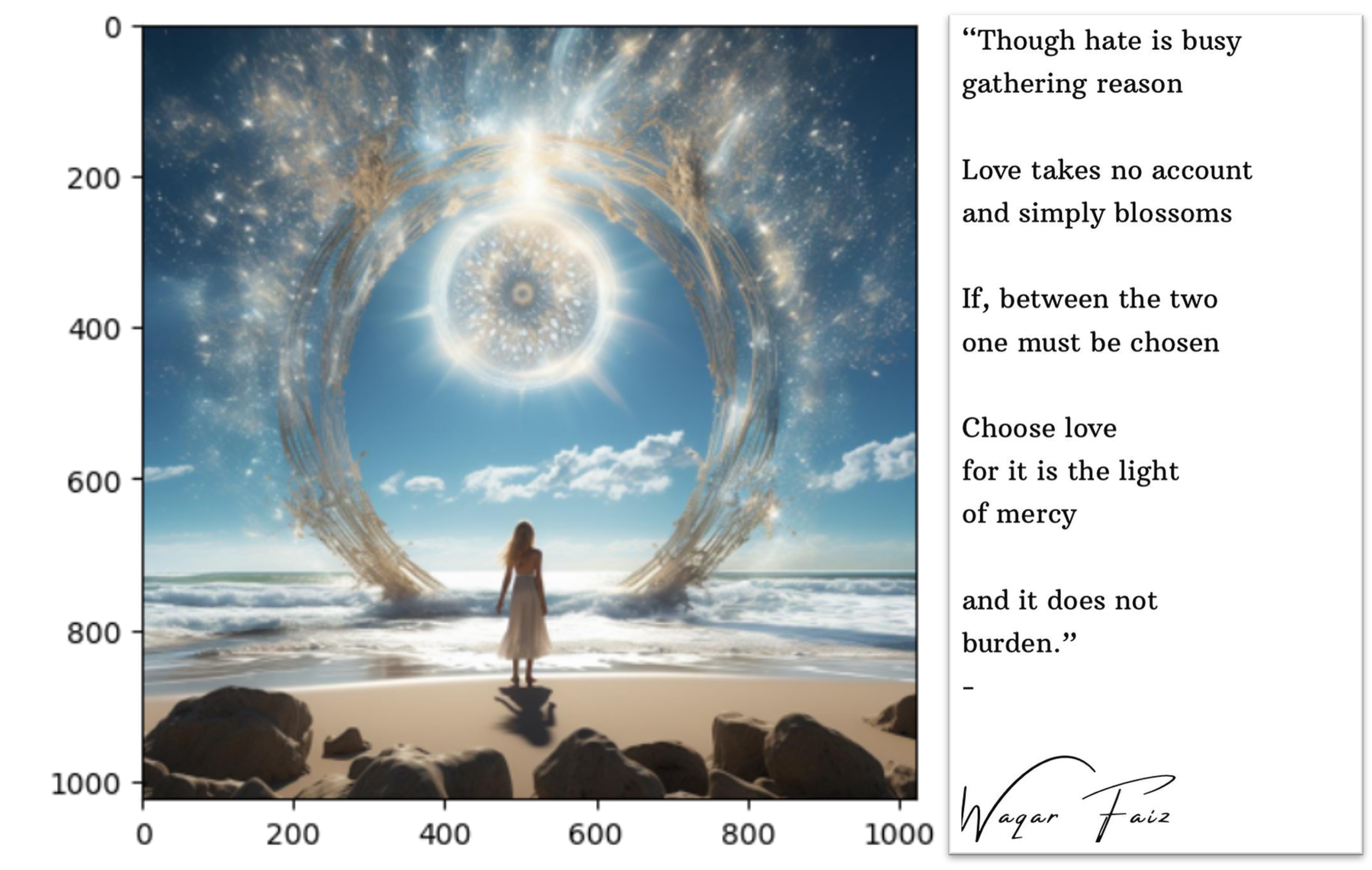}
    \caption{Instagram Image Example: The Left One Only Contains Visuality and The Right One Only Contains Textual Information}
    \label{fig:image annotation}
\end{figure*}

\subsection{Visual Analysis of Online Spirituality}
The visual culture prevalent on the web naturally connects with holistic spirituality, which emphasizes intuitive understanding and spiritual experience \citep{Noomen2011}. Therefore, images posted online serve as reflective mediums, offering insights into the mental health of individuals. They not only depict visual narratives but also convey underlying psychological states and circumstances, making them valuable for understanding and assessing mental well-being. \citeauthor{Tanhan2020} (2020) studied the biopsychosocial spirituality of Muslims using an online photovoice methodology. They advocated for the shaping of public and mental health professional training, especially addressing issues related to the public and mental health services for Muslim communities. \citeauthor{Xue2024} (2024) investigated the therapeutic potential of religion-related films by analyzing their distinctive emotional trajectories, through facial recognition techniques with YOLOv5, observing that the subtler emotional arcs foster introspection and personal growth. Also, images can be a tool serving as a sense of personal connection. Psychic practitioners draw on visual representations to convey their belonging to an exclusive cultural group using symbolic imagery to communicate shared knowledge and enhance their credibility within the psychic-spiritual community \citep{Ryan2012}. Sebek \citeyearpar{Sebek2019} studied the storytelling on Instagram and analyzed the transformative effects of spirituality when people share visual posts, such as selfies. \citeauthor{Trillo2021} (2021) also explored the visual contents on Instagram. They used a bottom-up taxonomy to classify images into categories and explored which visual repertoires are associated with users' value.

Generally, few studies integrate visual analysis techniques within the field of spirituality research \citep{Goldenfein2019}. This is due to two main challenges: firstly, automatic computer vision analysis techniques lack complete accuracy; secondly, hiring annotators is costly (cf. \citealp{Gilardi2023}), and annotating large datasets is both time-consuming and expensive. Our study intends to alleviate this.

\section{Data Creation}

\subsection{Dataset Sampling}
To collect data for this study, we developed a Python script utilizing the PRAW library to extract a substantial volume of posts from Reddit. The script adhered to Reddit's API usage guidelines, ensuring ethical and efficient data acquisition. We conducted a thorough examination of spirituality-related subreddits and selected ten for analysis: "Buddhism", "Christianity", "Enlightenment", "Meditation", "Mindfulness", "OpenChristian", "Soulnexus", "TrueChristian", "Spirituality", and "Taoism". The data was acquired on September 1st, 2024.

As to Instagram, people would like to share pictures and videos. Thus, inspired by the existing Python script, \textit{instagram-scraper}\footnote{https://github.com/meetmangukiya/instagram-scraper}, we extracted both images and their associated textual data including captions, hashtags, and mentions. In total, we collected data from 17 tags, including \#Buddhism, \#audiomeditation, \#awakening, \#christian, \#christainity, \#enlightenment, \#meditation, \#mindfulness, \#openchristian, \#soulnexus, \#spiritualawakening, \#spirituality, \#taoism, \#hindu, \#confucianism, \#islam, and \#truechristian. Considering the data analysis capability, we only collected 500 pictures for each tag. Due to limited data, some tags only contain less than 500 pictures, such as only 69 images with the tag \#soulnexus. Comparatively, the token numbers of the Instagram caption are much more than the subreddit posts.

For Quora data scraping, we drew inspiration from the existing Python script \textit{quora-plus-bypass}\footnote{https://github.com/NitinN77/quora-plus-bypass}. We got access to the Quora URLs specified in a text file, extracting both the accepted and suggested answers. There were 12 topics (Spirituality, Meditation, Enlightenment (spiritual), Buddhism, Christians, Taoism, Tao (Chinese philosophy), Yoga, Beliefs, Philosophy of Religion, Theology, and Atheism) in total involving four specific cultures: Christianity, Taoism, Buddhism, Atheism.

Eventually, we collected 6769 Instagram images as well as their captions, 3819 answers from Quora platform, and 4922 subreddit posts.

\subsection{Data Annotation}

\textit{Annotation Guidelines}

We invited five annotators \footnote{the fifth one is an external annotator for evaluation} whose major or minor is Computational Linguistics. Each annotator possesses a rich interdisciplinary background, having taken courses in social sciences and humanities in previous semesters. All annotators were trained to familiarize themselves with the annotation requirements and challenges of the dataset. We provided them with a detailed codebook (See Appendix 1. Annotation codebook), which included key term definitions and procedures, as well as illustrative examples. The annotation was conducted over one month on a part-time basis. This allowed annotators to thoroughly engage with the dataset, thereby enhancing the reliability of the annotations. All personal information has been deleted.

\textit{Annotating Textual Data}

For the present research, we use three distinct textual datasets: subreddit posts, Quora answers, and Instagram captions. Our annotators were tasked with a two-step annotation process. Initially, they determined whether a text was related to spirituality. If a text was spirituality-related, the next step involved classifying it into one of four types of connectedness. For Instagram captions, the annotators were also required to identify and record the language of the caption, which is useful for multilingual research of spirituality.

\textit{Annotating Images}

As part of our multimodal dataset, the annotation process extended beyond textual data to include images on Instagram. Annotators first judged whether an image could be classified as spiritual. For those images identified as conveying spiritual content, annotators were asked to determine the specific type of connectedness it represented. This classification drew upon the same categories used for textual data, as detailed in our codebook, ensuring a consistent approach across modalities. In addition to identifying spiritual content and its type, annotators were also instructed to note the language used in the images (e.g., memes or screenshots). Not all spiritual imagery includes textual language (see the left example in Figure \ref{fig:image annotation}) and some images only contain textual information (see the right example in Figure \ref{fig:image annotation}).

\section{Overview of the SACRED Dataset}

\subsection{Data Exploratory Statistics}

Among these, Quora exhibits the highest proportion of spirituality-related content at 53.94\%, significantly surpassing the other platforms. Instagram images show a relatively low engagement with spirituality-related content at 28.87\%, followed by Instagram captions at 21.27\%. Reddit features the lowest percentage of such content, standing at 17.76\%.

\begin{table}[h]
\centering
\caption{Percentage of Spirituality-related Content by Platform}
\begin{tabular}{lc}
\toprule
\textbf{Source} & \textbf{Spirituality-related (\%)} \\
\midrule
subreddit & 17.76\% \\
Instagram captions & 21.27\% \\
Instagram images & 28.87\% \\
Quora topics & 53.94\% \\
\bottomrule
\end{tabular}
\label{tab:spirituality_content}
\end{table}

\begin{figure}[ht]
    \centering
    \includegraphics[width=0.9\linewidth]{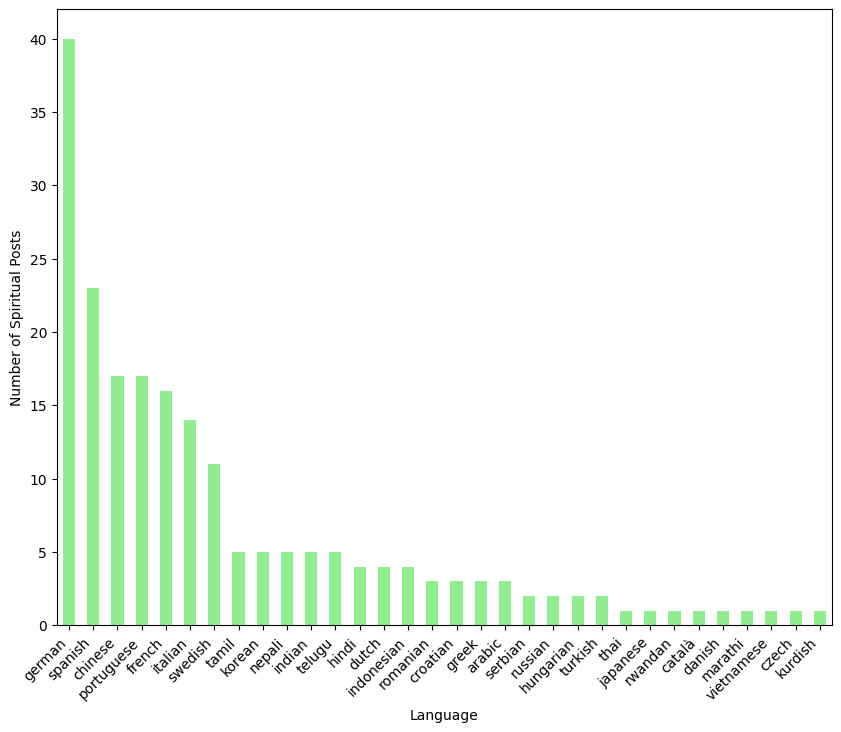}
    \caption{Language Distribution of Instagram Caption Excluding English}
    \label{fig:instagram caption}
\end{figure}

Given that our dataset is multi-lingual, we explored the language distribution of spiritual posts on Instagram, focusing specifically on posts excluding English. This sub-section provides a comprehensive overview of the linguistic diversity in spiritual communications on this platform.
The language distribution of Instagram captions, represented by green bars in Figure \ref{fig:instagram caption}, indicates a significant prevalence of German, Spanish, and Chinese.

\begin{figure}[ht]
    \centering
    \includegraphics[width=0.9\linewidth]{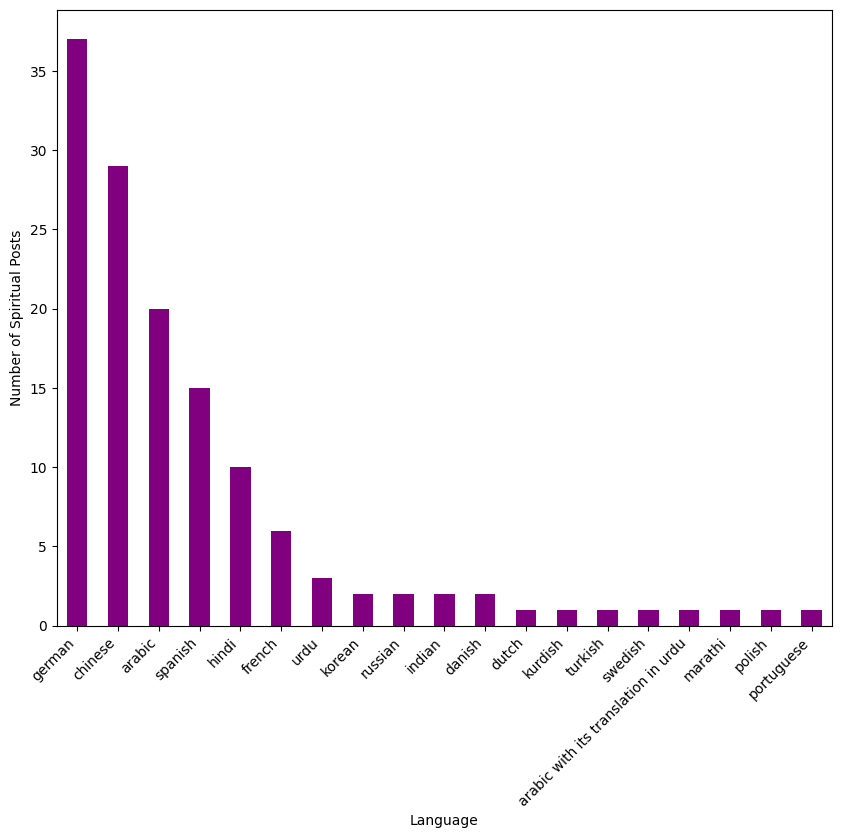}
    \caption{Language Distribution of Instagram Images Excluding English}
    \label{fig:instagram images}
\end{figure}

Conversely, the language distribution for text within Instagram images, shown by purple bars in Figure \ref{fig:instagram images}, presents a different pattern. German remains dominant, followed by Chinese and Arabic.

\subsection{A New Type of Connectedness}

A notable finding of data analysis is that 12 posts were identified as relevant to spirituality but did not align with any of the four predefined types of connectedness. We employed ChatGPT-o3 to analyze the content of these unclassifiable posts (the prompt is given in Appendix 6). This analysis led to the identification of a new type of connectedness.

\textbf{Connectedness to Art} There are some posts about creating or engaging with art that reflects spiritual themes, such as character drawings with spiritual significance or abstract art that evokes spiritual contemplation. They highlight a connection to spirituality through creativity and artistic expression. Also, some posts described authors' life changes by reading spiritual books like "The Untethered Soul" or "The Bible Tells Me So", which shows a connection to spirituality through the lens of literary influence and personal interpretation. In discussions with several experts in spirituality studies, one scholar suggested that connectedness to art could be understood as either connectedness to the self or to transcendence. Works of art, such as paintings or classical music, are created by artists (e.g., Vincent van Gogh) and resonates deeply with the audiences. Philosophically, this form of connection can be interpreted as an engagement with the artist's spirit or creative essence, which differs fundamentally from the inner spiritual purification typically achieved through meditation.

\section{Evaluation}

\subsection{Tasks}
To rigorously assess the efficacy of different computational approaches in classifying spirituality as well as connectedness types within our annotated datasets, we implemented a comparative evaluation framework. This framework included a variety of methods: a lexicon-based rule method, a fine-tuned BERT model, and LLMs for classifying the connectedness types of 193 images and 447 posts \footnote{We limited our evaluation to a smaller dataset because the small LLMs were tested locally on Ollama, and processing the entire dataset would have required several months of computation time. In addition, the outputs of small LLMs often failed to follow the given instructions, requiring extensive human post-processing of the annotation results.}. The selection of these images and posts was proportionally sampled according to the distribution of each hashtag to ensure representative coverage of different thematic categories.

\subsection{Experimental Setting}

We tested the performance of 13 types of models (11 LLMs and 2 traditional models). LLM prompts were attached in Appendix (see Appendix 7), and an introduction to each model is shown below.

\paragraph{Rule-based method} We employed rule-based methods to develop four dictionaries, each comprising keywords representative of a specific connectedness type, and subsequently performed classification based on these dictionaries.

\paragraph{Fine-Tuning BERT} We utilized the 
uncased version of the BERT model \citep{Devlin-BERT} and fine-tuned it on a selection (20\%: 875 rows in total, proportionally sampled from each platform) of the whole dataset.

\paragraph{LLaMA3.1-8B} We incorporated the LLaMA3.1 model with 8B parameters (text-only), which is a small, open-source, but prominent model of the LLaMA series \citep{Touvron-2023}.

\paragraph{Reka Flash-21B} It is a powerful 21B-parameter language models trained on 5T text tokens \citep{Ormazabal-2024}. We selected this model because 15\% of the pretraining data of Reka is multilingual involving 32 diverse languages.

\paragraph{GPT} We integrated three close-source GPT models \citep{Achiam-gpt}, including GPT-4o, GPT-4o-mini, and GPT-4.1-mini. The comparison is able to reveal the performance difference between smaller, distilled version and full-featured version.

\paragraph{Qwen2.5VL-7B} To test the visual classification ability, we utilized Qwen2.5VL model, which is a multimodal model with 7B parameters by Alibaba Group \citep{Qwen-2025}.

\paragraph{DeepSeek-V3-671B} DeepSeek-chat is built on Multi-Head Attention (MHA) block of a transformer and Mixture-of-Experts architecture \citep{DeepSeek-2025}. The MHA accelerate inference requiring far less KV cache. Also, Deepseek proposed fine-grained expert segmentation and shared expert isolation.

\paragraph{LLaVA-v1.6-13B} LLavA is "a multimodal model that combines a vision encoder and Vicuna for general-purpose visual and language understanding" \citep{Liu-LLAVA-2023}. This model performed well in computational social science, such as image captioning \citep{Sun-2025}.

\paragraph{Claude-Sonnet-4.5} Released on 29 September 2025, Claude-Sonnet-4.5 demonstrated significant advancements over its predecessors, particularly in handling complex agents and coding tasks. It represents Anthropic’s most advanced multimodal model available at the time of writing.
 
\paragraph{Gemini-2.5-Flash}  Gemini-2.5-Flash, which excels in multimodal understanding, such as the ability to process up to three hours of video content, is Google's most advanced multimodal model available at the time of writing \citep{Gemini-flash-2.5}.

\paragraph{Gemma-3-12B}  Gemma3 is a lightweight, open-source large language model by Google. It is built from the same research and technology used to create the Gemini model \citep{Gemma3-Techreport}.

\begin{table*}[ht]
\centering
\caption{Accuracy Performance Comparison Across Platforms and Methods}
\begin{tabular}{>{\centering\arraybackslash}m{2.2cm}ccccccc}
\toprule
\textbf{Model} & \multicolumn{2}{c}{\textbf{Reddit}} & \multicolumn{2}{c}{\textbf{Instagram Caption}} & \textbf{Instagram Image} & \multicolumn{2}{c}{\textbf{Quora}} \\
\cmidrule(r){2-3} \cmidrule(r){4-5} \cmidrule(r){6-6} \cmidrule(r){7-8}
 & \textbf{Zero-} & \textbf{Definition-} & \textbf{Zero-} & \textbf{Definition-} & \textbf{Definition-} & \textbf{Zero-} & \textbf{Definition-} \\
 & \textbf{shot} & \textbf{injected} & \textbf{shot} & \textbf{injected} & \textbf{injected} & \textbf{shot} & \textbf{injected} \\
\midrule
{Rule-based} & 22.67 &  & 20.27 & - & - & 10.74 & - \\
{Fine-tuning BERT} & 53.33 & - & 46.62 & - & - & 48.32 & - \\
{LLaMA3.1} & 58.00 & 66.00 & 45.27 & 54.05 & - & 53.02 & 52.35 \\
{Reka Core} & 52.00 & 55.33 & 43.00 & 45.33 & - & 31.00 & 28.33 \\
{DeepSeek-V3} & \textbf{72.00} & \textbf{78.67} & \textbf{77.70} & \textbf{77.70} & - & \textbf{73.15} & \textbf{79.19} \\
{GPT-4o} & 67.33 & 76.00 & 59.46 & 65.54 & 74.60 & 47.65 & 59.73 \\
{GPT-4o-mini} & 67.33 & 68.67 & 54.05 & 61.49 & 75.82 & 50.34 & 55.70 \\
{GPT-4.1-mini} & - & - & - & - & \textbf{76.04} & - & - \\
{Gemini-Flash} & - & - & - & - & 73.06 & - & - \\
{Sonnet-4.5} & - & - & - & - & 73.96 & - & - \\
{Qwen2.5VL-7b} & - & - & - & - & 68.31 & - & - \\
{LLaVA-13b} & - & - & - & - & 40.37 & - & - \\
{Gemma3-12b} & - & - & - & - & 45.03 & - & - \\
\bottomrule
\label{tab:Performance Comparison}
\end{tabular}
\end{table*}

\subsection{Main Results}

Our experiment evaluated the textual classification performance of the rule-based method, fine-tuned BERT, and a series of LLMs (see Table \ref{tab:Performance Comparison}). DeepSeek-V3 outperforms all other methods in classifying the connectedness types of posts. For instance, in the definition-injected setting, it achieves the highest scores on Quora Answers (79.19\%). Across the board, models that utilize the definition-injected setting show possible improved performance compared to their zero-shot counterparts. For example, LLaMA3.1's score on Instagram caption increases from 45.27\% to 54.05\% when definitions are injected. The traditional methods (rule-based and fine-tuning BERT) lag behind the advanced LLMs, which shows the potentials of LLMs in settings where contextual understanding is crucial. 

In the Instagram image classification task, GPT-4o-mini achieved the highest F1 score (63.99\%) (see Appendix 8), while GPT-4.1-mini attained the highest accuracy. These results indicate that the ChatGPT model series continues to outperform other models in the task of classifying visual spirituality connectedness. What surprised us was the performance of the Qwen2.5-VL model, which achieved an accuracy of 68.31\% despite being a relatively small model with 7 billion parameters, outperforming LLaVA-1.5 model with 13 billion parameters and Gemma3 model with 12 billion parameters. Besides, although Gemma-3 and Qwen-2.5-VL are smaller models, their outputs remain consistently well formatted as large models, such as GPT-4o and DeepSeek-V3.

Comparing the classification performance among platforms, models generally perform better on Reddit posts and Instagram captions than on Quora answers. Our error analysis revealed that Quora answers tend to be relatively long, which means that a single response may contain elements of multiple connectedness types, even if the overall focus is on one primary type. While human annotators can readily identify the dominant connectedness type, LLMs may be more influenced by local contextual cues (\citealp{Zeng2025}), leading to misclassifications.

\section{Discussion}

To control for variables, we did not enable reasoning in our experiments, even though we believe that doing so could have yielded higher performance scores (e.g., DeepSeek-R1\footnote{\url{https://api-docs.deepseek.com/news/news250528}} and GPT-5\footnote{\url{https://platform.openai.com/docs/guides/reasoning}}). Our findings revealed that while advanced models like DeepSeek-V3 and GPT-4o-mini exhibit superior performance, they are not yet fully reliable in classifying abstract concepts inherent in spirituality. Notably, the inclusion of a strict definition of spirituality (see Section \ref{spirituality_definition}) can help the models understand the task in some unclear posts, highlighting the importance of contextual information in processing complex tasks. Also, we observed an interesting phenomenon that GPT models sometimes generated the label "connectedness to transcendence" even though the prompt explicitly instructed them to choose from five predefined categories. We know the reason is some scholars use "connectedness to transcendence" in their studies (e.g., \citealp{Rahe-2024}). This suggests a limitation in the model’s ability to strictly adhere to prompt constraints (i.e., hallucination), even when the instruction was explicit.

Despite these contributions, our study has several limitations:

Our dataset, while multimodal, predominantly features English-language content. The language distribution analysis shows some diversity, but languages like German and Spanish are underrepresented. This limits the applicability of our findings to a global context where spirituality is expressed in myriad languages. In the further research, we will incorporate data from a wider array of platforms and cultural contexts to enhance the representatives of our dataset.

\section{Conclusion}

In this paper, we introduced \textbf{SACRED}, the first multimodal multilingual annotated dataset specifically designed for online spirituality research. Our work underscores the complexity of classifying abstract and deeply personal concepts using computational methods for communication sciences. While advanced LLMs show promise, they are not infallible and require careful contextualization and guidance. The \textbf{SACRED} dataset provides a valuable foundation for future research, but it also highlights the need for ongoing refinement of both computational tools and theoretical models.

\section{Acknowledgements}

This paper was inspired by the seminar “Language Technology for Social Media Analysis” held in 2023 by Martin Volk (1961-2025†). This work is dedicated to his memory, honoring his lifelong contributions to the field of computational linguistics.

\bibliographystyle{lrec2026-natbib}
\bibliography{lrec2026-example}

\section*{Appendices}

\textbf{1. Annotation codebook}

The topic of the document is spirituality. You are expected to annotate the connectedness type for all posts and images. In this research, we use a strict definition of spirituality:

\begin{quote}
    Spirituality is the pursuit and practice of experiences, beliefs, and values that influence and nurture the spirit, fostering personal growth, meaning, and a sense of connection to something greater than oneself.
\end{quote}

Connectedness Type:

(1) \textit{Connectedness to Nature} \par
Connectedness to nature refers to the deep sense of relationship that individuals feel with the natural world and understanding of humanity's place within the broader ecological system.

(2) \textit{Connectedness to the Self} \par
Connectedness to the self includes authenticity, inner harmony/inner peace, consciousness, self-knowledge and experiencing and searching for meaning in life.

(3) \textit{Connectedness to Others} \par
Connectedness to Others emphasizes empathy, compassion, and a sense of community, recognizing that interpersonal connections contribute to personal growth, a deeper understanding of oneself and others, and overall spiritual well-being.

(4) \textit{Connectedness to Transcendence} \par
Connectedness to Transcendence pertains to "something or someone beyond the human level, such as the universe, transcendent reality, a higher power or God".

\textbf{Note:} Meditation is not necessarily connectedness to self. If something superhuman appears in it, it also belongs to connectedness to transcendence.

I. Instagram Annotation

There are two types of Instagram data: text data (Caption) and image data (image).

(1) \textit{Text data annotation}

Text data annotation is to classify the content in the "Caption" column. \par
- First determine whether the caption is spirituality-related (whether it has an impact on the spirit); \par
- If it is spirituality-related, mark it as 1; otherwise, mark it as 0; \par
- If it is spirituality-related, check which connectedness type it is (if it does not belong to any of them, you can mark it as \textbf{None of them} or name it yourself, but please note that the labels you mark should be consistent throughout the entire annotation process). \par 
- You need to additionally mark the language of the content (i.e. you can use ChatGPT or copying the caption to Google/DeepL Translate).

(2) \textit{Image data annotation}

Image data annotators need to find the corresponding image folder based on the "Tag" column. \par
- Label the image as spirituality based on its content \par
- If it is spirituality-related content, check which connectedness type it is (if it does not belong to any of the above, you can mark it as \textbf{None of them} or name it yourself, but please make sure that the labels are consistent throughout the entire annotation process).\par
- You need to additionally mark the language on the image (i.e. you can use ChatGPT or copying the caption to Google/DeepL Translate).

II. Quora data annotation

The annotator needs to annotate the "Answer" column.
- If it is spirituality-related content, annotate 1; otherwise, annotate 0; \par
- If it is spirituality-related content, check which connectedness type it is (if it does not belong to any of them, you can mark it as \textbf{None of them} or name it yourself, but please make sure that the labels you annotate are consistent).

III. Reddit data annotation

The Reddit contents that need annotating is post. The annotator needs to annotate the "post" column. \par
- If it is spirituality-related content, annotate 1; otherwise, annotate 0; \par
- If it is spirituality-related content, check which connectedness type it is (if it does not belong to any of them, you can mark it as \textbf{None of them} or name it yourself, but please make sure that the labels you annotate are consistent).

\textbf{2. Agreement among annotators}

\begin{figure}[ht]
    \centering
    \includegraphics[width=1\linewidth]{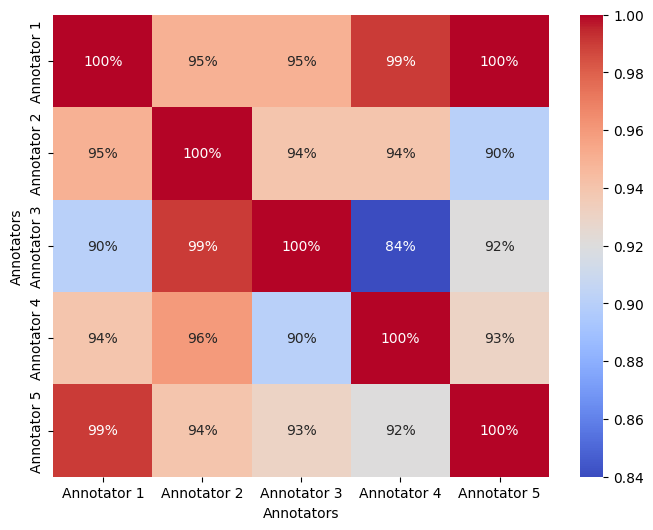}
    \caption{Heatmap of Annotation Agreement Among Annotators}
    \label{fig:heatmap}
\end{figure}

We engaged five annotators to independently review and annotate a dataset (\( \mathcal{N} \) = 100) previously curated by their peers. What they had to classify was the connectedness type. Then, we compiled the results and created a heatmap to visually represent the agreement and discrepancies among the annotations provided by each annotator. The figure \ref{fig:heatmap} shows a great agreement among five annotators.

\textbf{3. Connectedness types across different platforms}

This table displays the distribution of posts related to four types of connectedness across three social media platforms. The numbers represent the count of posts for each type of connectedness on each platform. 

\begin{table}[!htbp]
\centering
\caption{Connectedness Types Across Different Platforms}
\scriptsize
\begin{tabular}{>{\centering\arraybackslash}m{1.2cm}>{\centering\arraybackslash}m{0.9cm}>{\centering\arraybackslash}m{1.25cm}>{\centering\arraybackslash}m{1.25cm}>{\centering\arraybackslash}m{1.15cm}}
\toprule
\textbf{Type} & \textbf{Reddit Posts} & \textbf{Instagram Captions} & \textbf{Instagram Images} & \textbf{Quora Answers} \\
\midrule
Self & 542 & 766 & 1166 & 436 \\
Others & 28 & 146 & 89 & 109 \\
Nature & 22 & 80 & 83 & 12 \\
Transcendence & 282 & 444 & 873 & 1503 \\
\bottomrule
\end{tabular}
\label{sec:appendix_B}
\end{table}

\textbf{4. Unique representation of four types of connectedness}

Given that each connectedness type has its unique terms, we did a quantitative analysis, which focuses on quantifying the data by counting the frequency of each unique term associated with different types of connectedness. Initially, the texts were converted into its lemma form to standardize the vocabulary. Subsequently, terms specific to each connectedness category were identified by excluding words that appeared in more than one category. Each unique term was then counted to determine its frequency within its respective category. The results were displayed in horizontal bar charts.

The figure presents four separate horizontal bar charts, each representing the frequency of unique terms associated with four types of connectedness: Nature, Others, Self, and Transcendence. 

\begin{figure}[!ht]
    \centering
    \includegraphics[width=1\linewidth]{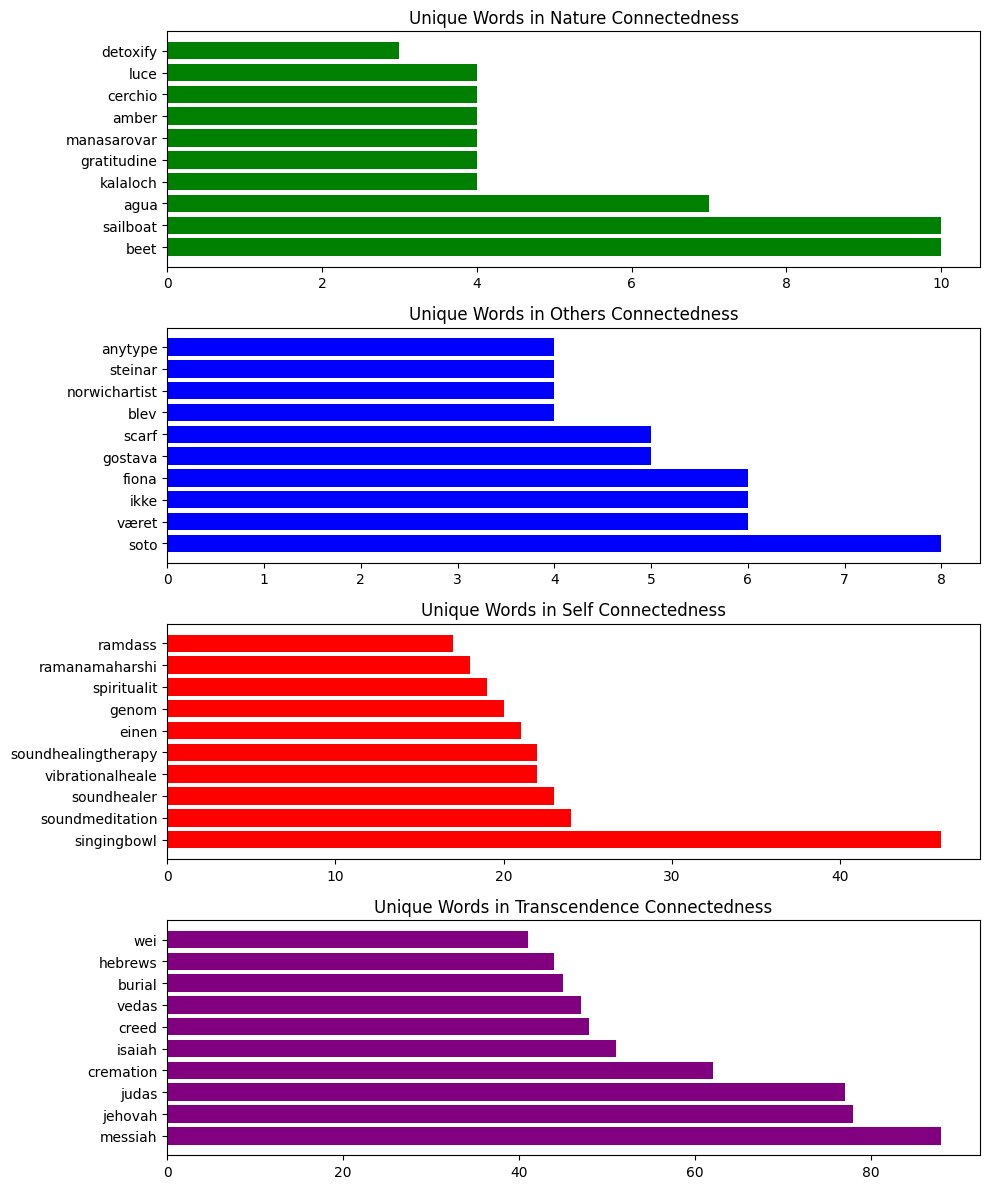}
    \caption{Unique Words in Four Types of Connectedness}
    \label{unique-term}
\end{figure}

The nature-related terms, such as "agua", "sailboat", "luce", and "manasarovar", suggest a focus on natural elements and experiences. As to connectedness to others, the feature words are "anytype", "steinar", and "norwichartist". Dominated by terms such as "ramdass", "spiritualiti", and "vibrationalheale", the chart of self connectedness suggests a focus on personal development, spirituality, and healing. The frequency of terms like "soundhealingtherapy" and "soundmeditation" highlights the sound-based therapeutic and meditative practices. The most populated words correlated to Transcendence like "messiah", "jehovah", and "vedas" reflect deep spiritual or religious beliefs and spiritual practices.

\textbf{5. Classification results of Fine-tuning BERT}

We employed the BERT model, specifically fine-tuned on our annotated dataset of Reddit posts. The fine-tuning was carried out 30 epochs, allowing the model sufficient iterations to adjust its parameters to the subtleties embedded in the dataset concerning different types of spiritual connectedness. The hyperparameter of learning rate is 2e-5, while we set the eps as 1e-8, which is a small constant added to prevent division by zero in the optimizer's calculations (AdamW optimizer), ensuring numerical stability. The figure below illustrates the trajectory of the model's accuracy over these epochs, showing the learning progress throughout the fine-tuning process. The model was uploaded on Huggingface (see \url{https://huggingface.co/QinghaoGuan/bert-spirituality-connectedness}).
\begin{figure}[!ht]
    \centering
    \includegraphics[width=0.8\linewidth]{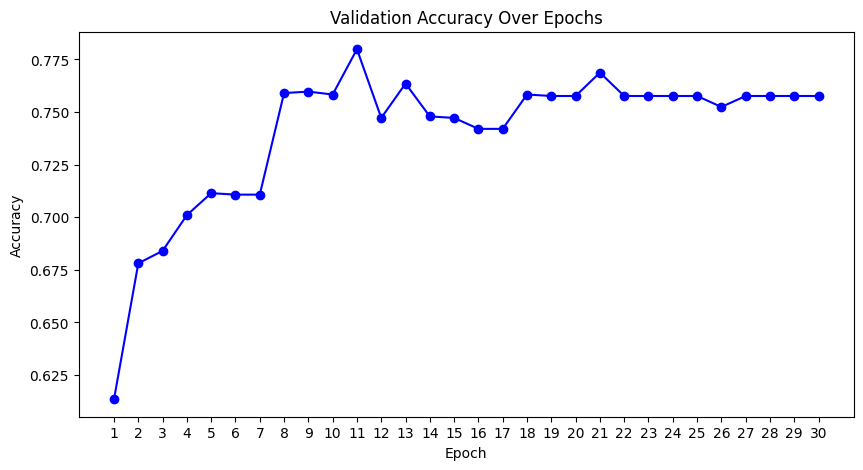}
    \caption{Validation Accuracy Curve over 30 Epochs}
    \label{fig:fine-tuning}
\end{figure}

\textbf{6. Prompt to guide LLMs to analyze topics of unclassifiable posts}

Here's our prompt to guide ChatGPT-4 in assisting with topic analysis:

\begin{quote}
   I am conducting research on spirituality by categorizing posts based on four types of connectedness: to self, to others, to nature, and to the transcendence. However, some posts do not fit neatly into these predefined categories. As an assistant, you are expected to analyze the topics of these unclassifiable posts. Based on your analysis, please identify and suggest an alternative type of connectedness that these posts might represent, which is distinct from the four established categories.
   
\end{quote}

\textbf{7. Prompts of zero-shot and definition-injected zero-shot}

Our prompts involve two settings, zero-shot and definition-injected zero-shot, and they are formulated as follows:

\paragraph{\textbf{\textit{Zero-shot}}} You are a human annotator and need to classify the sentence below into one of the following five labels:
"connectedness to self", "connectedness to others", "connectedness to nature", "connectedness to transcendence", or "hard to classify".
For each sentence, simply return self, others, nature, transcendence, or hard to classify.
Sentence: "\{post\}"

\paragraph{\textbf{\textit{Definition-injected zero-shot}}} You are a human annotator and need to classify the sentence below into one of the following five labels:
"connectedness to self", "connectedness to others", "connectedness to nature", "connectedness to transcendence", or "hard to classify".
Now I give you the definition of four types of connectedness. \{definitions\}. Based on the definitions above, classify the sentence "{post}" into one of five labels. For each sentence, simply return "self", "others", "nature", "transcendence", or "hard to classify".

\textbf{\textit{→ Definition given by LLMs}}

As mentioned in section \ref{spirituality_definition}, the definition of spirituality is changing all the time. Given that social scientists are also interested in LLMs' understanding of spirituality, we asked Reka model to conclude the definition of spirituality based on human annotation results. Here is the answer by the Reka model.
\begin{quote}
    Based on the task of inferring definitions for the four types of "connectedness", here are my interpretations: \par
    1. **Connectedness to Nature**: This refers to the sense of bond, harmony, and interdependence an individual feels with the natural world around them. It encompasses an emotional, spiritual, and sometimes physical relationship with ecosystems, flora, fauna, and the broader environmental context. This connectedness often involves a respect for and appreciation of natural processes and a feeling of belonging within the natural order. \par
    2. **Connectedness to Self**: This type of connectedness involves a deep understanding, acceptance, and integration of one's own identity, values, emotions, and thoughts. It's about having a strong sense of self-awareness and self-compassion, and feeling a cohesive and authentic sense of who one is. Connectedness to self also includes being in tune with one's needs, desires, and boundaries, and living in alignment with these internal truths. \par
    3. **Connectedness to Others**: This refers to the quality and depth of relationships and interactions an individual has with other people. It encompasses feelings of empathy, love, support, and mutual understanding with family, friends, colleagues, and even strangers. Connectedness to others involves a sense of community, belonging, and interdependence, and the ability to form and maintain meaningful bonds. \par
    4. **Connectedness to Transcendence**: This type of connectedness involves a sense of connection to something greater than oneself, which may be understood through spiritual, philosophical, or existential lenses. It can include feelings of awe, wonder, and reverence for the universe, a higher power, or the inherent meaning and purpose of life. Connectedness to transcendence often involves a sense of unity with all things, a feeling of being part of a larger cosmic or spiritual plan, and a quest for deeper understanding and insight beyond the material realm.
\end{quote}

\textbf{8. Precision, recall, and precision of image annotation by LLMs}

\begin{table}[htbp]
\centering
\scriptsize
\begin{tabular}{lccc}
\hline
\textbf{Model} & \textbf{Recall (\%)} & \textbf{Precision (\%)} & \textbf{F1 (\%)} \\
\hline
GPT-4o-mini    & \textbf{63.42} & \textbf{65.66} & \textbf{63.99} \\
GPT-4o         & 60.46 & 62.87 & 61.34 \\
GPT-4.1        & 48.42 & 59.68 & 49.83 \\
Gemini-2.5   & 55.07 & 63.79 & 56.99 \\
Claude-Sonnet  & 56.03 & 60.60 & 57.18 \\
LLaVA-1.5       & 34.83 & 47.44 & 38.96 \\
Qwen2.5-VL     & 61.94 & 55.76 & 54.74 \\
Gemma3     & 43.35 & 52.50 & 34.87 \\
\hline
\end{tabular}
\label{tab:llm_performance}
\end{table}

This table presents the classification performance of seven models on the task of categorizing images into different types of connectedness.
Overall, GPT-4o-mini achieved the highest overall performance, with a recall of 63.42\%, a precision of 65.66\%, and an F1-score of 63.99\%. GPT-4o followed closely, with slightly lower scores across all metrics (recall = 60.46\%, precision = 62.87\%, F1 = 61.34\%). This small performance gap suggests that while both GPT-4o and GPT-4o-mini are capable of handling complex visual-semantic classification tasks, the lightweight variant demonstrates slightly better generalization in this specific setting. Open-source small LLMs perform worse. Our findings show that scaling and fine-tuning strategies in recent GPT architectures have improved their visual-semantic reasoning capabilities, making them promising tools for computational spirituality and online religion research.

\end{document}